\documentclass[letterpaper, 10 pt, conference]{ieeeconf}  
\IEEEoverridecommandlockouts                                
\let\labelindent\relax
\usepackage[english]{babel}
\usepackage[T1]{fontenc}
\usepackage[ruled,vlined]{algorithm2e}
\usepackage{geometry}
\usepackage{amsmath}
\usepackage{graphicx}
\usepackage{amsfonts}
\usepackage{multirow}
\usepackage{enumitem}
\usepackage{subfigure}
\usepackage[utf8]{inputenc} 
\usepackage{hyperref}       
\usepackage{url}            
\usepackage{booktabs}       
\usepackage{nicefrac}       
\usepackage{microtype}      
\usepackage{gensymb}
\usepackage[table,xcdraw]{xcolor}
\usepackage{caption}
\usepackage{color}
\usepackage{wrapfig}
\usepackage{amssymb}
\usepackage{pifont}
\usepackage[table,xcdraw]{xcolor}
\usepackage{algpseudocode}
\usepackage{bm}
\usepackage{cite}
\usepackage{wrapfig}
\usepackage{stfloats}
\usepackage{dsfont}

\author{Qiang Wang, \textit{Student Member, IEEE}, Pablo Martinez Ulloa, Robert Burke, David Cordova Bulens, \\ and Stephen J. Redmond, \textit{Senior Member, IEEE}
\thanks{This publication has emanated from research conducted with the financial support of China Scholarship Council under grant number 202006540003 and of Science Foundation Ireland under grant numbers 17$/$FRL$/$4832 and SFI$/$12$/$RC$/$2289$\_$P2.}
\thanks{The authors are with the School of Electrical and Electronic Engineering, University College Dublin, Belfield, Dublin, D04 V1W8, Ireland (e-mail:\{qiang.wang, pablo.martinezulloa, robert.burke1\}@ucdconnect.ie; \{david.cordovabulens, stephen.redmond\}@ucd.ie).}
\thanks{\textit{Corresponding author: Stephen J. Redmond}.}
}

\title{\LARGE \bf \vspace{-15pt}
Robust Learning-Based Incipient Slip Detection using the PapillArray Optical Tactile Sensor for Improved Robotic Gripping}

\begin{document}

\maketitle
\thispagestyle{empty}
\pagestyle{empty}
\begin{abstract}
The ability to detect slip, particularly incipient slip, enables robotic systems to take corrective measures to prevent a grasped object from being dropped. Therefore, slip detection can enhance the overall security of robotic gripping. However, accurately detecting incipient slip remains a significant challenge. In this paper, we propose a novel learning-based approach to detect incipient slip using the PapillArray (Contactile, Australia) tactile sensor. The resulting model is highly effective in identifying patterns associated with incipient slip, achieving a detection success rate of 95.6\% when tested with an offline dataset. Furthermore, we introduce several data augmentation methods to enhance the robustness of our model. When transferring the trained model to a robotic gripping environment distinct from where the training data was collected, our model maintained robust performance, with a success rate of 96.8\%, providing timely feedback for stabilizing several practical gripping tasks. Our project website: \url{https://sites.google.com/view/incipient-slip-detection}.

\end{abstract}

\section{Introduction} \label{section-intro}
\subsection{Background}
Autonomous robots have yet to achieve human-like dexterity when performing gripping tasks, mainly due to a lack of satisfactory tactile perception and processing abilities. Studies have shown that even humans struggle with simple gripping tasks in the absence of tactile sensation \cite{human-no-tactile, human-no-tactile2}. The palm of the human hand contains $\sim$17,000 mechanoreceptors, i.e., specialized nerve endings that respond to mechanical stimuli such as deformation, pressure, and displacement \cite{human-hand-property, incipient-slip-detection-survey}. These receptors play a crucial role in sensing and relaying tactile information to the nervous system \cite{human-hand}, allowing humans to adjust their grip in real-time to account for slipperiness and other factors. Building on these insights, researchers have designed tactile sensors replicating part of human hand sensing capabilities and explored slip detection techniques using these sensors to enhance robotic manipulation performance \cite{papillarray-incipient-heba, gross-slip-1, gross-slip-2}. 

\subsubsection{Types of slip}
The two main types of slip are gross slip and incipient slip. Gross slip refers to the occurrence of slip across the entire contact surface, where the relative motion between the gripper or tactile sensor and the gripped object is typically observable at a macro level \cite{finger1, finger2}. On the other hand, incipient slip refers to the initial stage of slip, when parts of the contact surface slip while others remain stuck \cite{finger1, finger2}. For example, when an object is held by elastic fingertips, and an external force is applied to the object in a direction tangential to the contact surface, some parts of the fingertips will stretch while others will compress, causing incipient slip at the periphery of the contact surface while the central part remains stuck. As the applied force increases, the slip will finally spread across the entire contact surface, leading to gross slip. Throughout the incipient slip phase, there may not be any observable relative motion between the object and the finger.   
\subsubsection{Slip detection and challenges}
Previous studies have proposed techniques to detect gross slip and apply corrective measures when the slip is detected to prevent objects from dropping out of the grasp \cite{gross-slip-1, gross-slip-2, gross-slip-3}. Detecting gross slip may not always be a wise strategy, as it occurs when the entire contact has already started slipping. On the other hand, detecting incipient slip can provide an early warning of an impending and more dangerous gross slip, allowing corrective measures to be applied earlier, and increasing the likelihood of maintaining a safe grip. However, detecting incipient slip is not trivial because it requires the contact interface of the sensor to possess adequate elasticity, enabling one part to undergo sufficient and detectable deformation, resulting in slip, while the other part remains stuck. Furthermore, validating incipient slip can be challenging since it is not generally associated with macro-level relative movement between the sensor/finger and the object. To verify the occurrence of incipient slip, researchers commonly utilize a camera to monitor the contact surface; by examining the camera images, they can visually confirm the presence of incipient slip events \cite{modify-tactip-incipient, papillarray-incipient-pablo}.  However, this method of relying on cameras may not be feasible in real-world situations, such as when gripping everyday objects.

\begin{figure}[t]
\begin{center}
    \centerline{
    \subfigure[]{\label{fig:sensor}
                            \includegraphics[width=0.281\columnwidth]{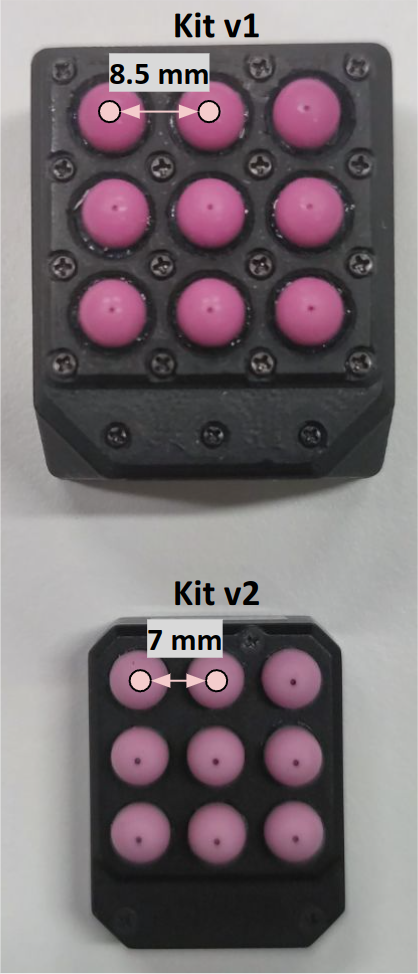}}
        \hspace{0mm}
    \subfigure[]{\label{fig:gripping-rig}
                            \includegraphics[width=0.655\columnwidth]{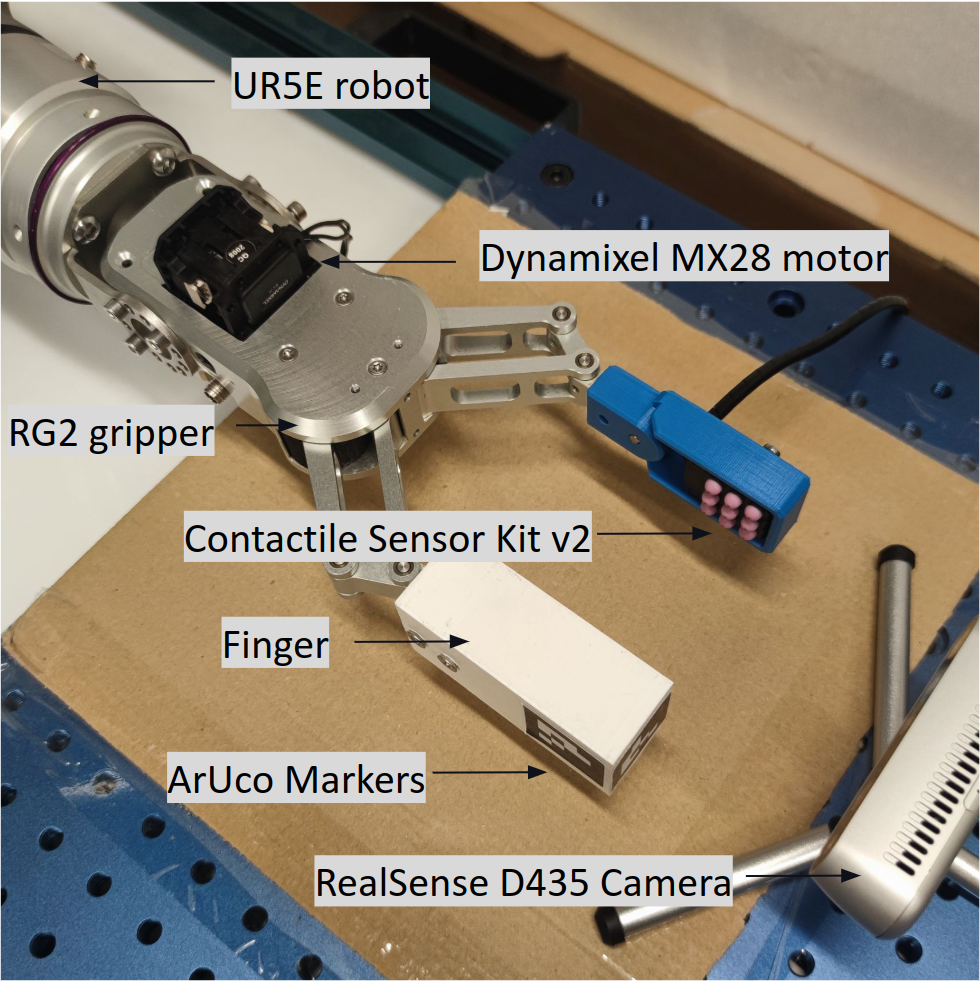}}
                            }
\end{center}
         \caption{(a) Illustration of two versions of the Contactile PapillArray Dev Kit sensor that were used in this study: the upper shows version 1 (v1) and the bottom shows v2. (b) Illustration of robotic gripping rig used to deploy our trained model for online evaluation.}
\label{fig-rigs}
\vspace{-1.0em}
\end{figure}


\subsection{Our contribution} \label{subsec-contri}
Our study presents a new technique for detecting incipient slip using the PapillArray (Contactile, Australia) tactile sensor. This sensor features a square array of nine elastic silicone pillars with varying unloaded heights, promoting different normal forces on the pillars when pressed against a surface. This design enhances the likelihood of inducing incipient slip on shorter pillars when a tangential force is applied.


We utilized deep neural networks (NN) to develop our incipient slip detection algorithm, where we made novel use of the data gathered in a previous study \cite{papillarray-incipient-pablo} to construct the dataset for training and evaluating the NN. The primary objective of the NN was to classify inputs into two distinct categories: \textit{incipient slip} and \textit{other}, functioning as a binary classifier; \textit{other} refers to all others states that are not incipient slip, such as gross slip or being stationary. Furthermore, the tactile data at hand is presented in the form of a uniformly-sampled time series. Therefore, to effectively capture the serial nature of the data, we utilize a recurrent neural network (RNN) \cite{rnn}. The inclusion of historical data in a NN model has the potential to enhance its performance in real-time prediction tasks, as it enables the capture of temporal patterns and dependencies, leading to more robust and accurate forecasts \cite{rnn}. We also propose several data augmentation methods designed to enhance the performance and robustness of our trained model, making it robust to environmental confounders.



\section{Related work} \label{Related work}


Similar to the approach we will take in this paper, the approach proposed in \cite{gross-slip-1} treats slip detection as a classification task; the authors employed a support vector machine \cite{svm} to detect slip using the velocity of embedded pins on the inner surface of a TacTip camera-based tactile sensor \cite{tactip}. Labels of the training data are assigned manually based on the alignment of pin velocities. In a more recent study \cite{modify-tactip-incipient}, the authors modified the TacTip sensor used in \cite{gross-slip-1} by introducing raised fingerprint-like ridges, decreasing skin thickness, and increasing pin spacing to reduce mechanical coupling between ridges and to create the traction differential and facilitating the shear displacement required for the occurrence of incipient slip. This is similar to the behavior seen on the human finger pad when sheared against an object, thus allowing the sensor to experience incipient slip. They used an external camera to monitor the contact in real-time for data labeling, and then employed a convolutional neural network \cite{cnn} as a binary classifier to detect incipient slip.

The GelSight technology is another camera-based tactile sensing system that uses an elastic body to establish a contact with an object, with the built-in camera recording the resulting deformation to obtain tactile data \cite{gelsight}. An approach was introduced in \cite{gelsight-detect-1} for detecting incipient slip using the GelSight sensor. This method determines the degree of incipient slip by analyzing the inhomogeneity of the displacement field, which is quantified in terms of entropy. More recently, a more advanced version of the GelSight technology, called GelSlim, was proposed in \cite{gelsight-detect-3}; it employed the deviation of the deformation field from a 2D planar rigid displacement field to determine slip.\vspace{0mm}


Compared to camera-based tactile sensors, the distributed optical sensor used in our work, the PapillArray, is less complex in terms of instrumentation\cite{papillarray-instrument}. It offers several advantages over other sensor designs, including size, temporal resolution, and compliance. A heuristic algorithm that employs the PapillArray tactile sensor to detect incipient slip is proposed in \cite{papillarray-incipient-heba}. The approach is based on the observation that incipient slip happens when some sensor pillars stop deflecting at the same rate as the contacted object is moving in the sensor's frame of reference. Precisely,  this approach detects slip by evaluating the tangential velocity drop with respect to a reference pillar, which is the pillar under the highest normal force (usually the center). In the case of rotational movements, with the center of rotation at the center pillar, the algorithm cannot detect any slip since no movement can be detected in the center pillar. This heuristic approach is further improved in \cite{papillarray-incipient-pablo} to account for rotational slips, detecting the deceleration of each pillar by comparing it to its own recent maximum velocity, and then it checks if other pillars are still in motion to confirm that the deceleration indicates an incipient slip. However, these methods may not be applicable when dealing with deformable or non-planar surfaces, or when only a subset of the pillars makes contact with the object. In such cases, establishing a dependable reference pillar to represent the object's movement in \cite{papillarray-incipient-heba} becomes challenging; in \cite{papillarray-incipient-pablo}, it is difficult to determine whether the deceleration of pillars is caused by slip or by the shape of the object's surface. 

In our work, we are motivated to take a learning-based approach in developing a dedicated incipient slip detection algorithm, where we propose domain adaptation techniques to enhance the robustness of our trained model, enabling it to effectively detect incipient slip for more realistic objects and contacts, overcoming the challenges outlined above.

\section{Materials and methods} \label{Materials and methods}
\subsection{Hardware}
\subsubsection{Contactile sensor}
Our study employed the commercial PapillArray sensor from Contactile\footnote{\url{https://contactile.com/}}, depicted in Fig.~\ref{fig:sensor}, which is based on the concept described in \cite{papillarray-concept}. The sensor outputs the real-time $x-y-z$ force data experienced by each pillar at a high sampling rate of 1,000 Hz. Our training data was collected using the Dev Kit v1, while for the online evaluation of our trained model, we used the Dev Kit v2. Dev Kit v2 and Dev Kit v1 differ in size and the pillar Shore hardness.
\subsubsection{Robotic gripping rig}
Fig.~\ref{fig:gripping-rig} displays the rig used in our study for the gripping task. The rig features a specialized two-finger gripper (RG2, OnRobot, Germany) with a blue adapter fixed to one of its fingers. This adapter serves to couple the Contactile PapillArray Dev Kit v2 sensor to the gripper finger. A white 3D-printed cuboid is used to extend another finger, matching the length of the finger equipped with the sensor. Moreover, a couple of ArUco markers are attached to this extended cuboid to track the gripper's pose. We replaced the original motor of the RG2 gripper with a stepper motor (MX-28, Dynamixel, US) to achieve high-frequency interruptible control of the gripper. The modified gripper was mounted on a six-axis robot arm (UR5e, Universal Robots, Denmark).

\subsection{Data preparation} \label{subset: dataset-preparation}
\subsubsection{Collect slip data and annotate slip events for individual pillars} \label{subsubsec:collect-slip-data}
Our training dataset is sourced from \cite{papillarray-incipient-pablo}. In brief, the training data was acquired using a six-degree-of-freedom hexapod robot (H-820, Physik Instrumente, Germany) with the Contactile PapillArray Dev Kit v1 sensor mounted on the top. A transplant acrylic plate is fixed above the sensor on a T-slot frame and a video camera (Logitech Streamcam, Logitech, Switzerland) is positioned above the acrylic plate to capture videos of the contact between the sensor and the plate. During the data collection, the hexapod pushes the sensor vertically against the acrylic place and then moves it laterally to induce a slip. The horizontal movement could be a translation, a rotation, or a combination of both. A total of 200 data sequences were collected, covering a range of compression levels, hexapod movement velocities, and movement directions. The recorded videos are processed using the Matlab Computer Vision Toolbox (MathWorks, USA) to track the pillar tip position. The tangential pillar tip velocity is then used to label the slip state (gross slip or not gross slip) of individual pillars.


\subsubsection{Collect control data} \label{control-group}
\begin{figure}[h]
\vspace{0.5em}
     \centering
         \includegraphics[width=0.95\columnwidth]{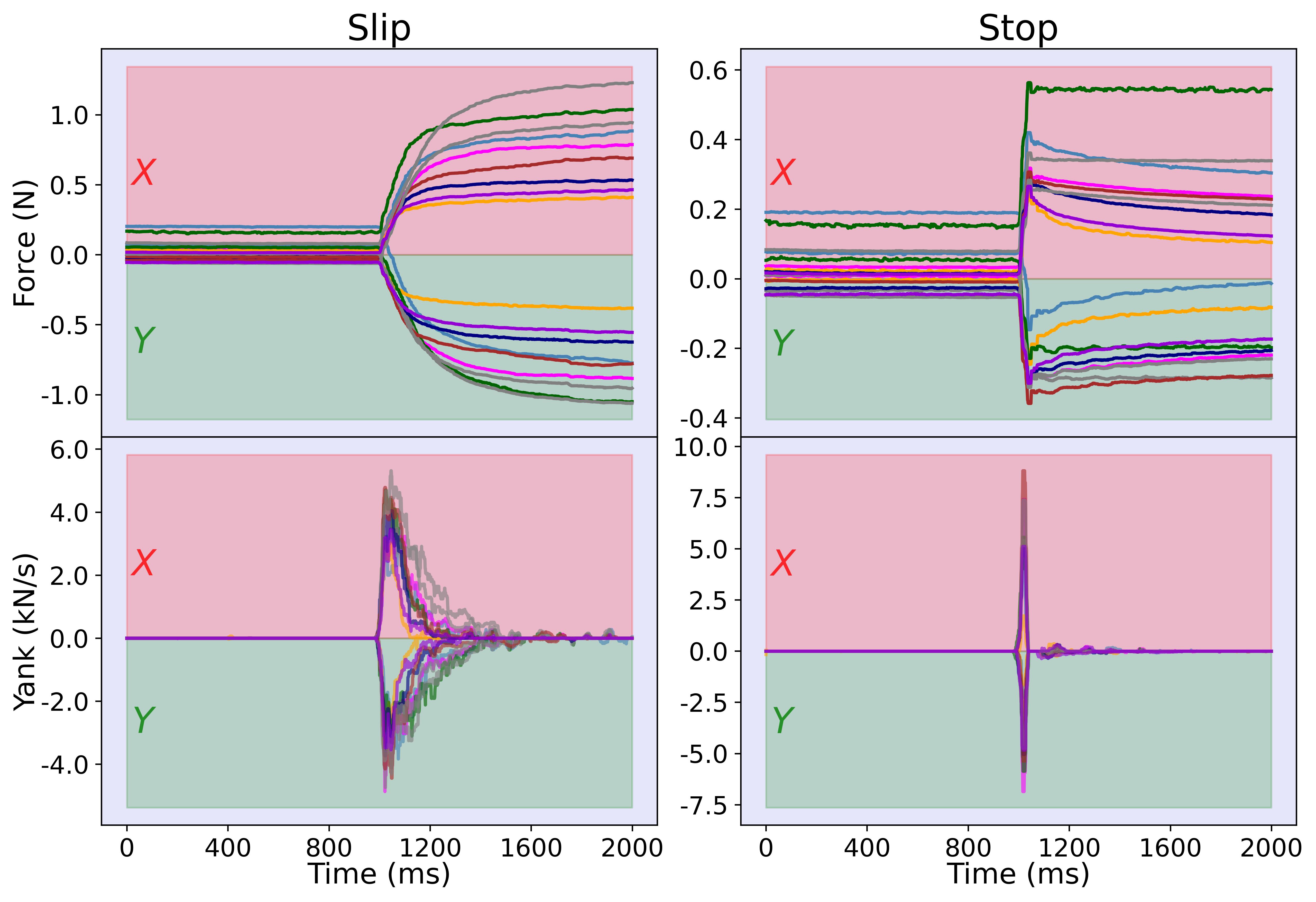}
         \caption{Changes in force and yank (rate of change of force) over time between slip and stop events. The X area (red) and Y area (green) represent the respective components in the $x$ and $y$ directions. The compression of the tallest pillar in these two examples is 1.3 mm, and the movement direction is at a 45$^{\circ}$ angle in the $x-y$ directions, with a speed of 8 mm.s$^{-1}$.} 
          \label{fig:control-group}
\vspace{-1.0em}
\end{figure}

When the sensor is compressed against a flat surface and moved laterally, the tangential velocity measured by each pillar will increase at first, as the sensor starts deforming, before reaching a peak velocity and subsequently decreasing its speed when the pillar stops deforming (Fig.~\ref{fig:control-group}). If a pillar stops deforming because it is undergoing incipient slip, at least one other pillar will still be deforming laterally; this is observed by an asynchronous decrease of the tangential velocity of the nine pillars (Fig.~\ref{fig:control-group} - Slip). However, if the object stops moving before any slip occurs, the tangential velocity magnitude of the nine pillars decreases almost simultaneously (Fig.~\ref{fig:control-group} - Stop).

Since the stop events display similar temporal feature to slip events, we collected an additional dataset specifically focusing on stop events, consisting of a total of 28 data sequences. We label the data points in these sequences as \textit{other}. By incorporating this dataset, the NN is less likely to misclassify between incipient slip and \textit{other}, thereby improving the accuracy and reliability of the NN. The data collection process was similar to that of the slip events, except that the hexapod's movement was abruptly halted before any slip occurred. Further details on this process can be found in \cite{papillarray-incipient-pablo}.

\subsubsection{Annotate the incipient slip} \label{annotate-incipient-slip}
Based on the definition of incipient slip provided in Section~\ref{section-intro}, we annotate the incipient slip in the dataset as follows: we consider that incipient slip has occurred when at least one pillar slips with respect to the contact surface, while at least one other pillar remains stationary with respect to the contact surface. In other words, we start annotating incipient slip from the moment the first slip occurs on any pillar, and this interval continues until the time when all nine pillars have slipped. The slip label of each pillar is obtained as described in Section~\ref{subsubsec:collect-slip-data}. It should be noted that when annotating incipient slip in the rotational data, we only consider the outer eight pillars. This is because the rotational movement is centered around the central pillar, which never slips by our definition (remains in the same location on the contact area), for our data set.


\begin{figure*}[ht]
\begin{center}
    \centerline{
    \subfigure[]{\label{fig:rnn}
                            \includegraphics[width=0.55\linewidth]{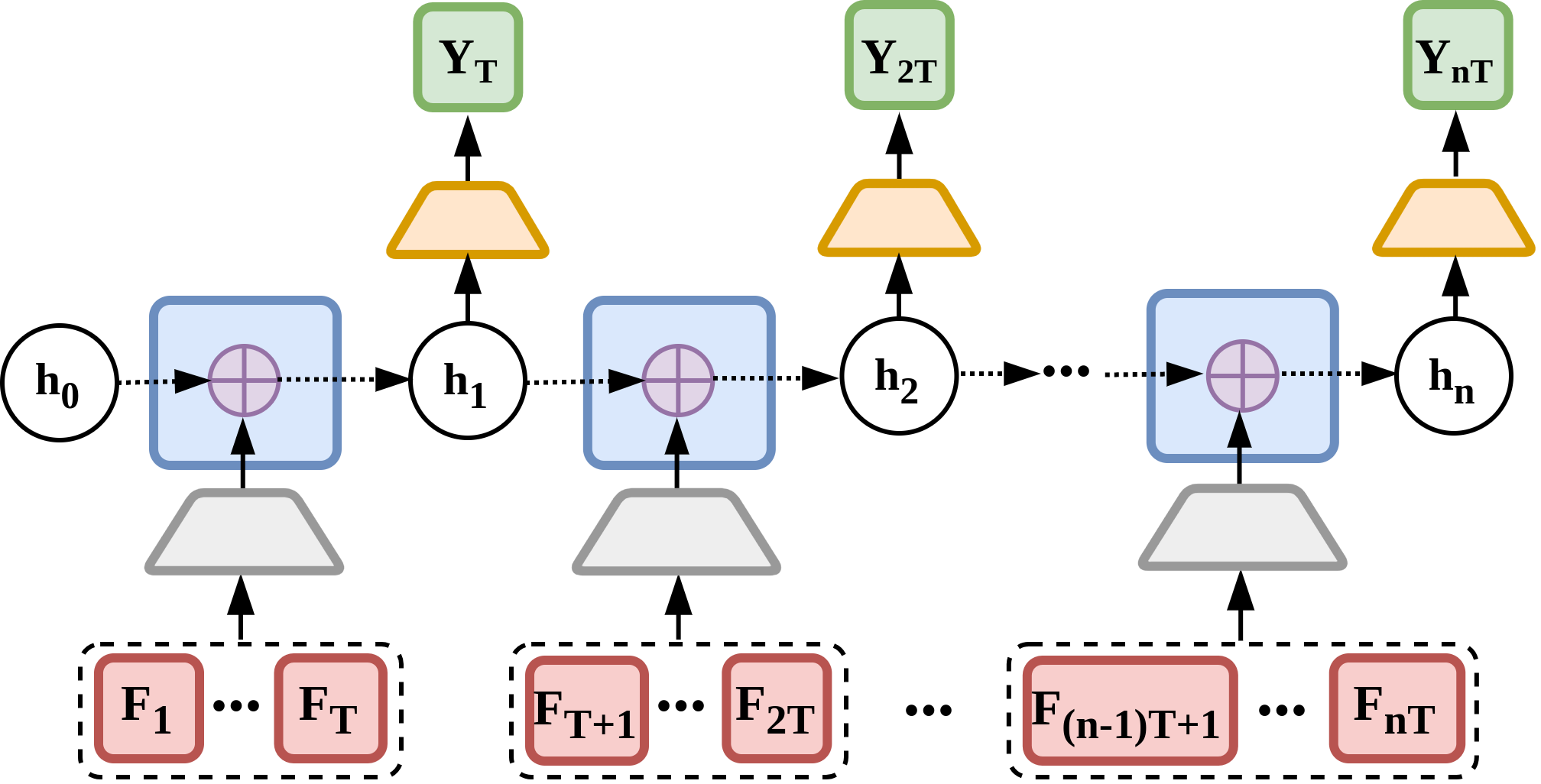}}
        \hspace{10mm}
    \subfigure[]{\label{fig:ensemble}
                            \includegraphics[width=0.163\linewidth]{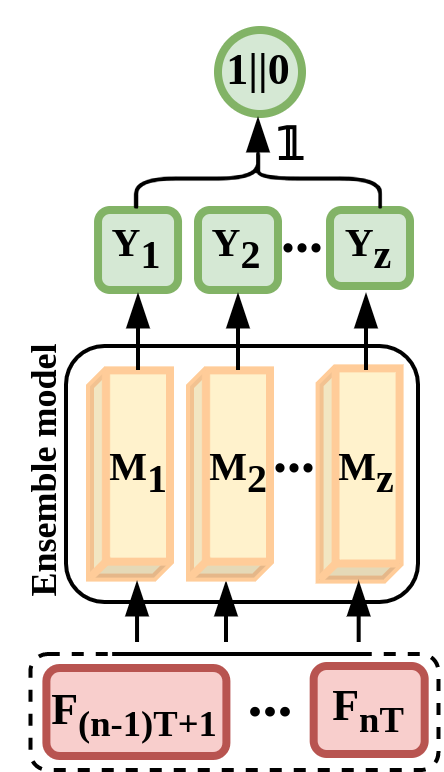}}
                            }
\end{center}
 \caption{(a) Illustration of how sequence data is input into a model to generate classifications for each window. The input is represented by the red rectangle, which undergoes window reshaping, denoted by the dotted rounded rectangles. $T$ represents the number of data points in one data sequence, and in our case $T=40$ samples at 1,000 Hz. The transparent circles represent the hidden states; the initial hidden layer, $h_{0}$, is initialized as a vector of zeros with the same shape as the hidden state of the GRU cells. The encoder is shown as grey trapezoids connected to blue GRU cells, with purple circles representing recurrent computations. The orange trapezoids represent the fully connected estimator, and the green items represent the output estimations. (b) Illustration the aggregation process when the ensemble model is deployed, using $Z=5$ classifiers in the emsemble.}
  \label{fig:networks}
\vspace{-0.5em}
\end{figure*}

\subsubsection{Refine data sequence} \label{subsubsec:modify-data-sequence}
The sensor output exhibits variance due to noise and sporadically produces glitches that deviate significantly from the mean value, displaying sudden extreme highs or lows. To address these issues, we apply a median filter with a window size of 21 samples on the raw sensor signal, which is sampled at 1,000 Hz.

We divided the raw data sequence into non-overlapping windows, with each window containing 40 samples. This division reduced the data rate to 25 Hz. This was done for practical limitations in the hardware and software of our system. More precisely, the maximum refresh rate of our gripper servo is $\sim$62~Hz, and the computation rate of our classifier is $\sim$40~Hz. Moreover, it is worth noting that reliable gripping does not necessarily require a high sampling frequency. Indeed, humans have a reaction time of approximately 80-120 ms (equivalent to 8.3-12.5 Hz) \cite{human-tactile-speed}, enabling us to perform most everyday gripping tasks effectively. 

Finally, we only consider the $x-y$ forces on the pillars as input in NN training, while excluding the $z$ force. During the data collection process, when the hexapod moves tangentially to induce slip, it remains stationary in the $z$ direction. As a result, we assume that the $z$ force does not play a significant role in detecting incipient slip in our case. It should be acknowledged that in real-world scenarios, the normal force can provide valuable information for humans to detect slip, and it is likely to vary appreciably for different gripping objectives. Therefore, another reason for excluding the $z$ force is to prevent the NN from incorrectly learning that the $z$ force remains relatively stable during slip events, as occurs in our data set.




\subsection{Training data augmentation} 
\subsubsection{Data augmentation by rotational symmetry} \label{aug1}
During the data collection process, the sensor is placed at the origin of the world coordinate frame. Its horizontal surface is parallel to the $x-y$ plane of the world frame of reference, and the side edges align with the $x-y$ axis directions. Hence we use a rotation transformation to augment the data; intuitively, it can be understood as rotating the initial position of the sensor around the $z$ axis by a random angle. For each data point in a sequence, we perform the following mathematical calculations:
\begin{equation}
\label{eq_rot_obj}
    \begin{bmatrix}F_{x'} \\ F_{y'} \end{bmatrix} = \begin{bmatrix}\cos(\theta) & -\sin(\theta) \\ \sin(\theta) & \cos(\theta) \end{bmatrix}\cdot\begin{bmatrix}F_{x} \\ F_{y} \end{bmatrix}, \theta \in[0,2\pi),
\end{equation}
where $F_{x}$ and $F_{y}$ represent the force values along the original $x-y$ axis, and $F_{x'}$ and $F_{y'}$ are the augmented force values after virtual rotation of the sensor by a randomly sampled angle, $\theta$, from a uniform distribution of $[0, 2\pi)$.


\subsubsection{Advanced data augmentation for domain adaptation} \label{aug2}
The data used in our study was collected under idealized conditions, where a hexapod robot was used to compress the sensor against a flat surface and move laterally in a controled manner. In this setup, the force was nearly perpendicular or parallel to the contact surface and the movement speed is nearly constant. However, in real-world robotic gripping, the conditions are expected to be quite different from this idealized setup, and the performance of the model trained on such data is expected to be poor. We identify several issues that may arise when transferring the model trained on idealized data to real-world gripping scenarios, and we propose a range of advanced data augmentation methods to address these issues in the following paragraphs. These methods are designed to generate synthetic data that mimics the real-world variability of the gripping: 


\begin{itemize}[leftmargin=*]
    \item \textit{\textbf{\textcolor{red}{Issue: }}The slipping velocity in real-world robotic gripping is not constant, as it is influenced by various factors such as gravity, friction, and the shape of the object being gripped. However, during the data collection process, the hexapod induces slip at a constant velocity}. \textbf{\textcolor{blue}{Remedy: }} We employ random sampling to sample a percentage of data points from the raw data sequence, thereby generating a new data sequence. And we maintain the frequency of the new sequence at the same rate as the raw sequence (1,000 Hz). This approach can simulate velocity variations to mimic real-world gripping scenarios, as it changes the magnitude differences of some temporally adjacent data points while keeping the time interval unchanged.

    \item \textit{\textbf{\textcolor{red}{Issue: }}In some gripping scenarios, a portion of the sensor pillars may not be in contact with the object. For instance, this can occur when employing sensors to grip an object with a rounded surface or when gripping an object smaller than the sensor's contact area}. \textbf{\textcolor{blue}{Remedy: }}To simulate an unloaded pillar, we substitute a number of pillar data sequences with zero sequences. Noise is then added to make the generated sequence resemble a realistic sensor signal. The noise is derived from a normal distribution with a mean of 0.0 N and a standard deviation of 0.001 N.

    \item    \textit{\textbf{\textcolor{red}{Issue: }}Unlike with the hexapod, the force generated by the gripper may not be perfectly perpendicular to the $x-y$ plane of the sensor frame of reference, and the force leading to slip may not be perfectly in this plane. This can occur when the gripped object is not flat or the mechanical linkage of the gripper flexes when applying force to the object}. \textbf{\textcolor{blue}{Remedy: }}First, we sampled nine individual pillar sequences from raw sensor sequences with different sensor compression levels and hexapod movement types, and then combined them to form a new sensor sequence. Secondly, we scaled (scale factor ranging from 0.2 to 2.0) the magnitude of values for a number of pillar sequences. Lastly, we randomly permuted the position (by pillar index) of a nine-pillar sequence. Employing these techniques can encourage the NN capture a broader and more comprehensive pattern of incipient slip (see Section~\ref{annotate-incipient-slip}), rather than only learning the limited pattern introduced by the hexapod.
    
\end{itemize}

\subsection{Neural networks}
The key decision making component of our incipient slip detection approach is a binary classifier. Initially, we trained a NN capable of estimating the probability of incipient slip for each time point in a sequence. Next, we set a threshold to convert the continuous probability into a binary output. To enhance the accuracy of the classifier, we used an ensemble technique that trains multiple independent classifiers concurrently and aggregates their output probabilities to produce the final decision (shown in Fig.~\ref{fig:ensemble}).

\subsubsection{Architecture}
Fig.~\ref{fig:rnn} illustrates the process of inputting a data sequence into the NN and obtaining the corresponding slip classification. The modified data sequence, as explained in Section~\ref{subsubsec:modify-data-sequence}, is input into an encoder. Subsequently, the encoder output is passed to a specific type of RNN called a gated recurrent unit (GRU)~\cite{gru}. In our approach, we utilize a single layer of GRU for each propagation step, and we refer to it as a GRU cell. The hidden output from the GRU cell is generated as a combination of the current input and historical information. Moreover, an estimator is included that takes the hidden layer output from the GRU cell and converts it into a probability estimation. The ground truth label of each window is determined by the label of the last sample in the window.

\subsubsection{Training}
The ensemble model consists of $Z$ ($Z=5$ in our case) independently trained classifier models. During each training iteration of each classifier model, a subset comprising a proportion of $\lambda$ sequences ($\lambda=40\%$ in our case) is randomly sampled with replacement from the entire training set and used for NN training. The final layer of the estimator utilizes a two-class softmax activation function, with its outputs interpreted as probabilities for the occurrence of incipient slip and \textit{other}. Our chosen loss function is binary cross-entropy. 




\subsubsection{Decision making}
We aggregate the output probability from each classifier model in the ensemble to convert the continuous probability to binary prediction:
\begin{equation}
f:=\mathds{1}\left[\frac{\sum_{z=1}^{Z}M_{z}(x=[F_{(n-1)T+1},\cdot\cdot\cdot,F_{nT}])}{Z} >P_{th}\right],
\label{eq:indicator}
\end{equation}
where $\mathds{1}[\cdot]$ is an indicator function, $M_{z}$ donates the $z^{th}$ classifier model in the ensemble,  $x$ donates the input vector, and $P_{th}$ denotes the probability threshold, which is 50\% in our work. $Z$ donates the number of classifiers in the ensemble model.

\begin{figure}[h]
\begin{center}
    \centerline{
    \subfigure[]{\label{fig:offline-prediction}
                            \includegraphics[width=0.97\linewidth]{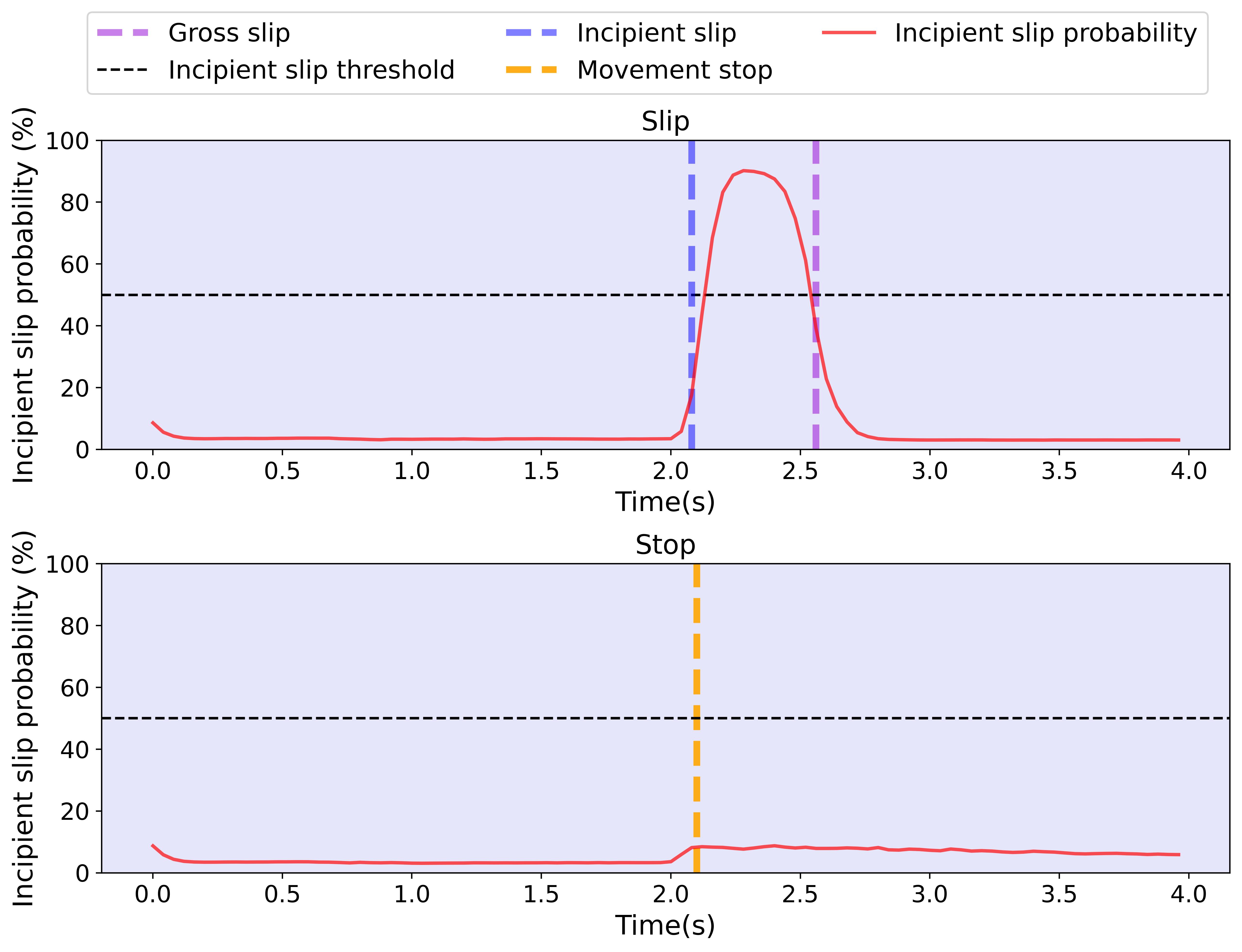}}}
    \centerline{             
    \subfigure[]{\label{fig:offline-conf}
                            \includegraphics[width=0.44\linewidth]{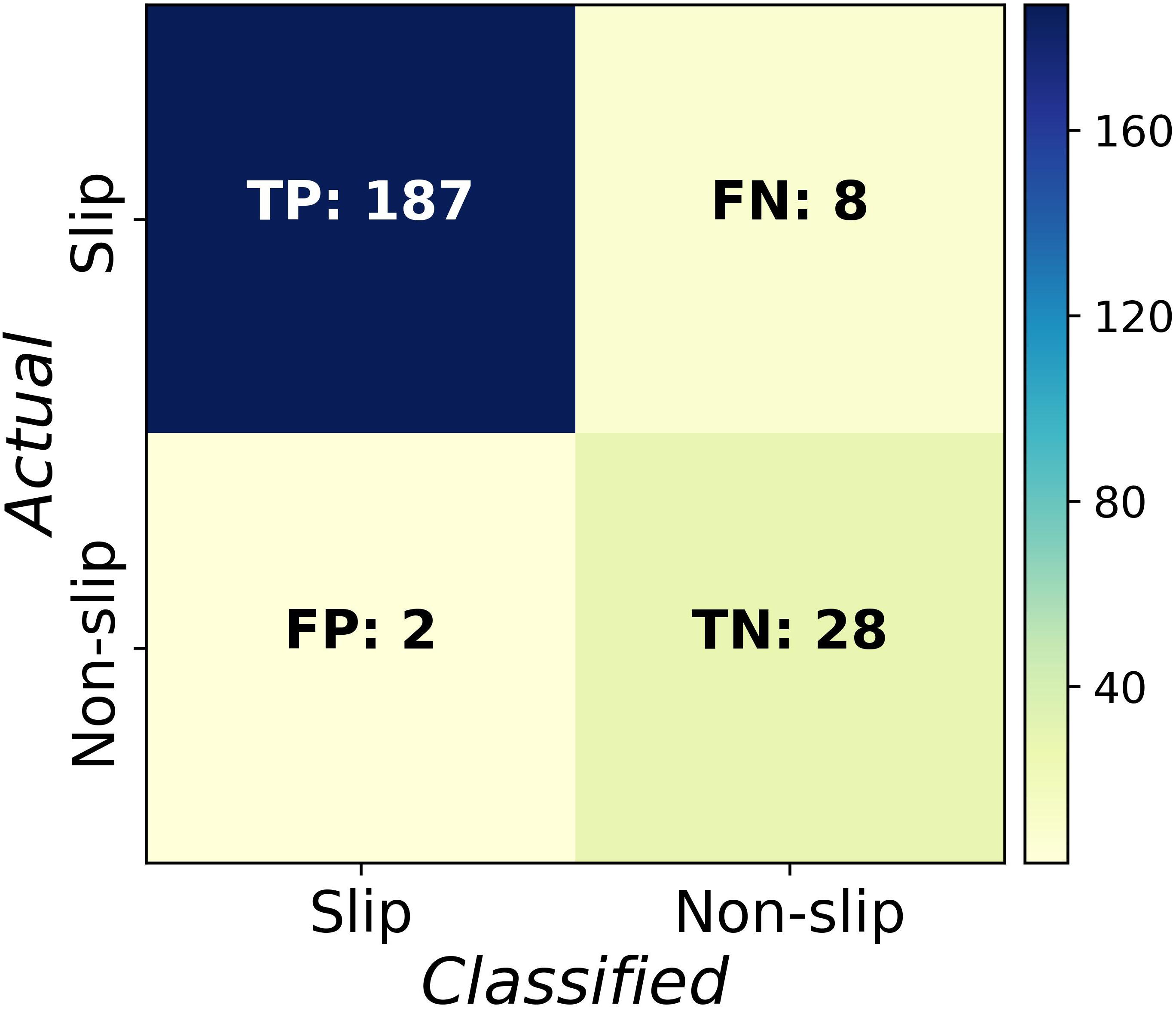}}
    \hspace{-8pt}
    \subfigure[]{\label{fig:offline-latency}
                            \includegraphics[width=0.52\linewidth]{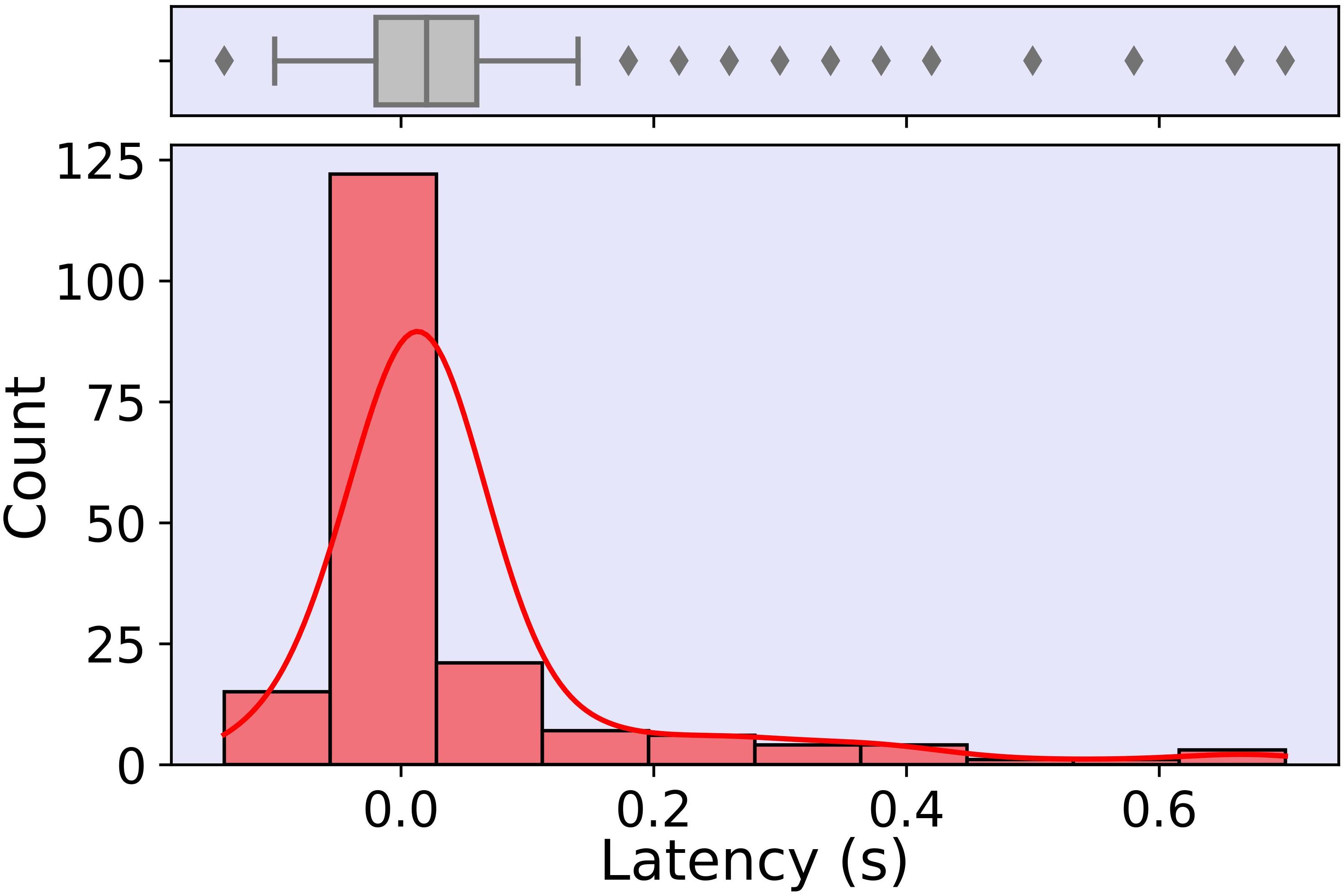}}}
       
\end{center}        
\caption{The offline evaluation results. (a) Illustration of two examples of successfully classified sequences, one corresponding to slip event and the other to stop event. (b) Illustration of the confusion matrix for the augmented test set, where TP represent successful detection of incipient slip, FN represents incipient slip is detected in a sequence of stop event, FP represents nothing is detected in a sequence of slip event, and TN represents the absence of detection in stop events, i.e. successful classification. (c) Illustration of the latency for successful TP cases in predicting incipient slip. This figure was created with reference to the moment when the first of the nine pillars slips (the ground truth for the onset of incipient slip).}
\label{fig:offline-results}
\vspace{-1.0em}
\end{figure}

\section{Experiments and results} \label{Experiments and results}
We first explicitly display our method's high success rate in detecting incipient slip, including offline and online scenarios. Then, we illustrate the practical benefits of our approach by showcasing its ability to stabilize an insecure robotic grasp in a number of practical gripping tasks.

\subsection{Offline evaluation} \label{subsec-offline-evaluation}
The entire dataset is randomly split into two subsets: a training set ($\sim$80\% of the entire dataset, comprising 160 data sequences of slip event and 23 data sequences of stop event) for model training, and a test set ($\sim$20\% of the entire dataset, consisting of 40 data sequences of slip event and 5 data sequences of stop event) for model evaluation. Both subsets are expanded through the symmetry-based augmentation method described in Section~\ref{aug1}, resulting in a five-fold increase in the size of the training set and test set.

Fig.~\ref{fig:offline-prediction} displays two examples comparing the incipient slip detection results over slip and stop events. As observed, the algorithm's confidence in labeling incipient slip increases rapidly as incipient slip starts and decreases as it progresses toward gross slip. In comparison, the probability in the stop case fluctuates slightly but remains well below the threshold.

We define an incipient slip detection as successful if it occurs within a 0.3 second window preceding the true labeled time point of incipient slip (to accommodate the error of the ground truth) and prior to the occurrence of the gross slip. For the stop event, a successful estimation is defined as a classification of the entire sequence as \textit{other}.


Fig.~\ref{fig:offline-conf} presents the confusion matrix, displaying the final classification results over the entire test set; our algorithm achieves an overall success rate of $\sim$95.6\%. The results also demonstrate its effectiveness in differentiating between the slip and stop events; this indicates that our algorithm is not simply detecting the changes in the force and yank of the pillars, as mentioned earlier in Section~\ref{control-group}. 

Our algorithm can effectively detect incipient slip in its early stages. 
In Fig.~\ref{fig:offline-latency}, we present the latency between the moment of incipient slip detected by the algorithm and the ground truth onset of incipient slip. It is evident that, on average, incipient slip can be detected within 10~ms of its initiation.


\begin{figure}[ht]
     \centering
         \includegraphics[width=0.99\linewidth]{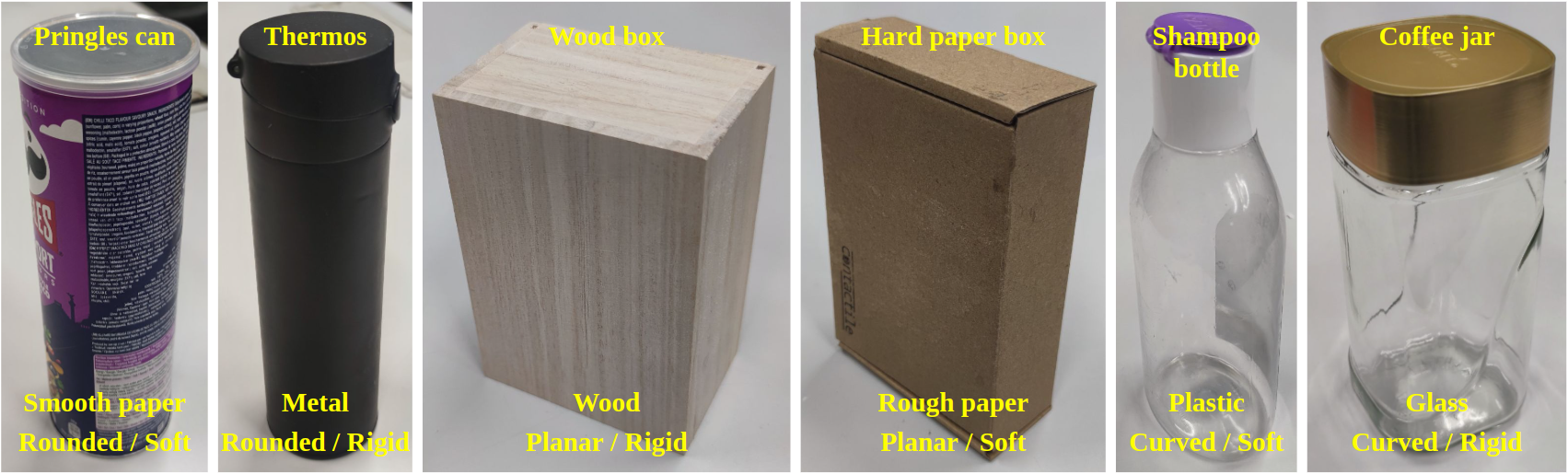}
         \caption{Everyday objects for our online gripping task.}
          \label{fig:objects}
\vspace{-1.0em}
\end{figure}

\begin{figure*}[h]
\begin{center}
    \centerline{
    \subfigure[]{\label{fig:online-confusion}
                            \includegraphics[width=0.243\linewidth]{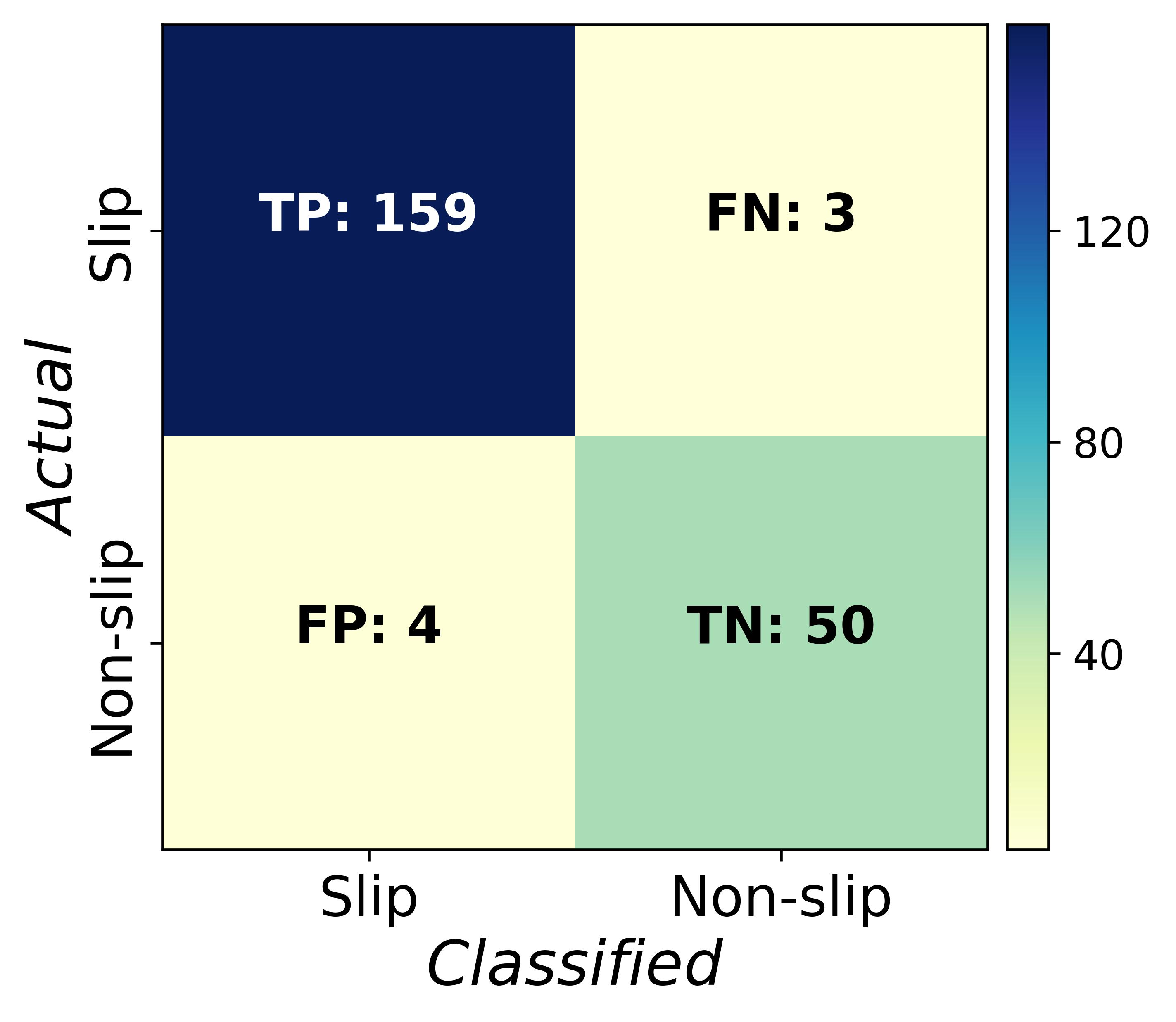}}
        \hspace{0mm}
    \subfigure[]{\label{fig:online-latency}
                            \includegraphics[width=0.738\linewidth]{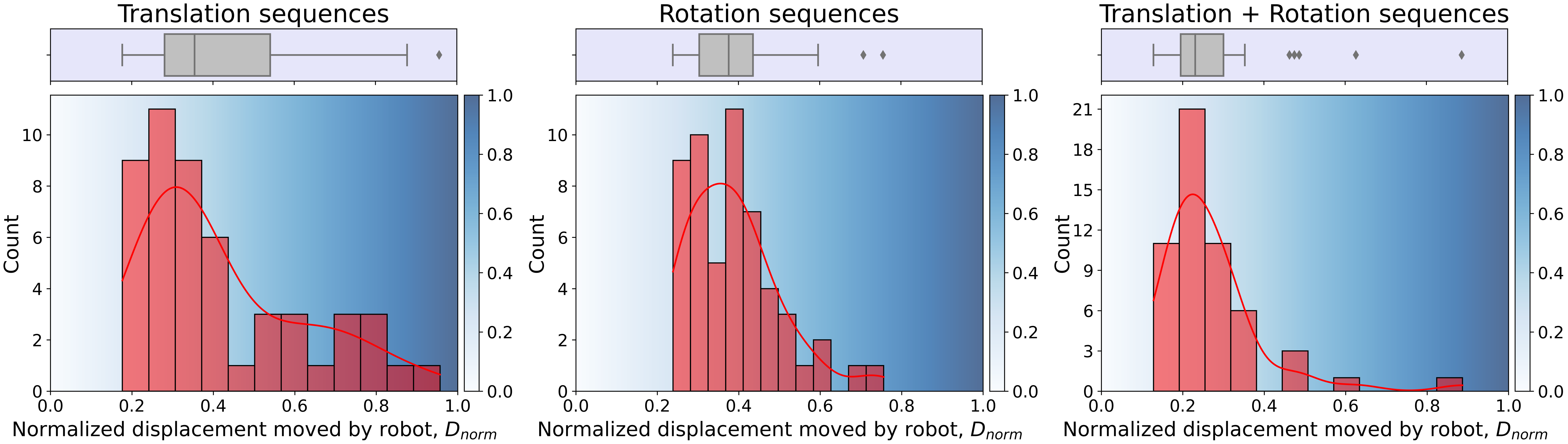}}
                            }
\end{center}
\caption{(a) Illustration of a confusion matrix for the classification. (b) Illustration of our validation of incipient slip in online testing. The $x$ axis in each sub-figure represents the normalized robot translational/angular displacement, calculated as $D_{norm} = D / D_{th}$, where $D$ is the displacement before normalizing and $D_{th}$ (2~mm for translation and 2$^{\circ}$ for rotation) is an assumption that indicates the reasonable minimum distance at which the gross slip must have happened or is likely to happen. To calculate $D_{norm}$ for the compound Translation + Rotation sequences, we calculate the $D_{norm}$ for both the translation and rotation displacements respectively and then take their mean value. $D_{norm}$ facilitates the intuitive visualization of when incipient slip is detected during movement of the robot and also facilitates comparison across three different types of movement. The blue gradient fill shows increasing confidence of gross slip, starting at 0.0 when the robot starts moving and reaching 1.0 at $D_{th}$.}
\label{fig:online1}
\vspace{-1.0em}
\end{figure*}

\subsection{Online evaluation}
In the online evaluation stage, we utilized the full data set for training the final deployed model. Again, to increase the amount of training data, we applied both symmetry-based (see Section~\ref{aug1}) and advanced data augmentation (see Section~\ref{aug2}) techniques, resulting in a five-fold increase in data amount (1140 data sequences). 


The online evaluation was performed on six everyday objects, depicted in Fig.~\ref{fig:objects}. We include objects of varying surface materials, curvatures, and hardness to ensure a broad range of conditions are represented in our results.

\subsubsection{Validating incipient slip detections} \label{subsubsec-online-success-rate}
We cannot easily validate incipient slip occurrences for everyday objects as we cannot independently monitor individual pillar contacts. Hence, we choose to perform the online evaluation based on following well-founded assumptions. The incipient slip detection is considered successful if it can be detected at any time-point between the time when the robot's movement begins ($T_{m}$) and the time when gross slip occurs ($T_{g}$); the criterion for determining the occurrence of gross slip has been arbitrarily defined as the occurrence of relative translational movement greater than 2 mm or relative rotational movement exceeding 2$^{\circ}$ between the object and the robot's frame of reference.

To induce a slip, the gripper first grips the object with a constant force. Then the robot moves the gripper downwards towards a rigid and stationary table surface, eliciting the slip between the sensor attached to the gripper tip and the object. In each trial, the gripping force is selected from a range of 8~N to 30~N. The robot movement can be either translational, rotational or a combination of translational and rotational. The velocity ($v$) and acceleration ($a$) of the robot movement have three different levels: low ($v = 4\text{ mm.s}^{-1}$, $a = 10\text{ mm.s}^{-2}$), medium ($v = 10\text{ mm.s}^{-1}$, $a = 50\text{ mm.s}^{-2}$), and high ($v = 40\text{ mm.s}^{-1}$, $a = 100\text{ mm.s}^{-2}$). All robot movements were performed using the built-in \textit{movel} function of the UR script. The tool center position and orientation are obtained using the built-in \textit{getl} function of the UR robot. This function employs forward kinematics calculations based on the read joint angles. 


In accordance with the offline evaluation, control trails are also conducted here for each $v$ and $a$ combination and movement type. The purpose is to validate that the identified behavior is indeed the incipient slip, rather the event with similar pattern like the stop event we mentioned above. The control data involves lifting the robot arm while maintaining a secure grip using a pre-determined grip force that is sufficient to prevent any slippage. As a result, when lifting an object, the pillars in contact undergo downward deformation due to the force of gravity; subsequently, once the object is securely held by the gripper and remains relatively motionless, these pillars will remain stationary. Here, for the sake of convenient explanation, we will also refer to this event as stop, and we label the sequence as \textit{other}. To ensure a fair experiment, we add extra weight to lightweight objects to enhance their downward motion when being lifted, aiming to make the pattern of the output data sequence more like a slip event. In total, our experiment consisted of 216 trials, including 162 sequences of slip event (6 objects $\times$ 3 movements $\times$ 3 forces $\times$ 3 velocity/acceleration combinations) and 54 sequences of stop event (6 objects $\times$ 3 movements $\times$ 1 force $\times$ 3 velocity/acceleration combinations).


Fig.~\ref{fig:online1} illustrates the final validation results. Fig.~\ref{fig:online-confusion} shows a confusion matrix, highlighting the high success rate ($\sim$96.8\%) of our method in detecting incipient slip and its ability to differentiate between slip and stop events. Fig.~\ref{fig:online-latency} demonstrates that our algorithm can detect incipient slip almost immediately upon the initiation of the movement that induces slip, with a normalized displacement $D_{norm}$ range of 0.2 - 0.4, within which the incipient slip can be detected (refer to the caption for the definition of $D_{norm}$). These results provide comprehensive validation of the effectiveness of our approach in detecting incipient slipping in real-world gripping tasks.



\subsubsection{Ablation study}
\begin{figure}[h]
  \centering
  \includegraphics[width=0.496\linewidth]{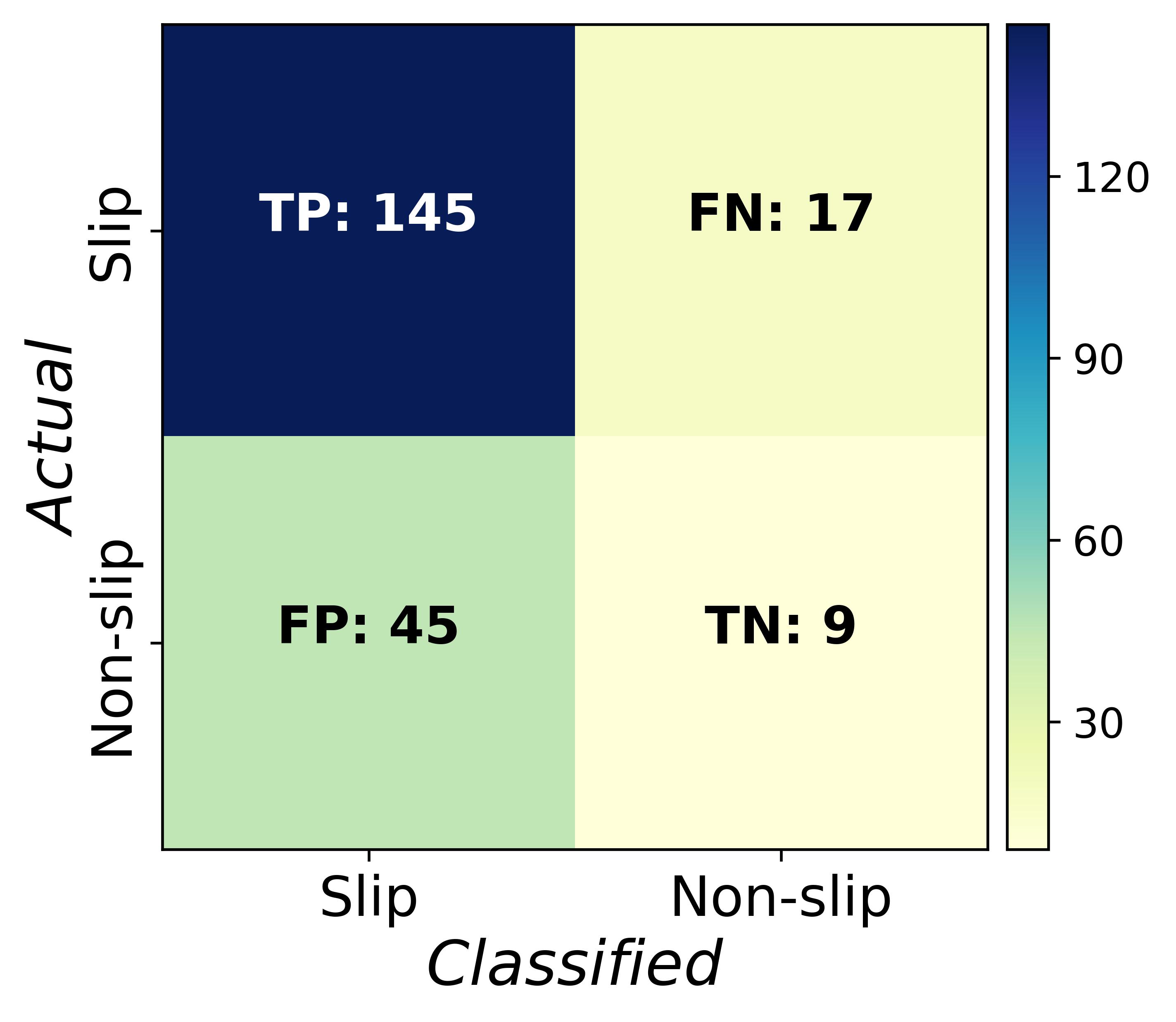}
  \caption{The ablation study results are to be compared with those presented in Fig.~\ref{fig:online-confusion}. The model here is trained without using data augmentation of the training data set leading to an increased number of misclassifications.}
  \label{fig:ablation-study}
\vspace{-1.0em}
\end{figure}

This study aims to showcase the effectiveness of our advanced augmentation method in bridging the domain gap between the idealized data collected with the hexapod and more realistic data encountered with the robotic gripper. To accomplish this, we employed the model training approach described in Section~\ref{subsec-offline-evaluation}. However, instead of splitting the data into separate train and test sets, we trained the model using the entire dataset here, given the different objective. Subsequently, we conducted online gripping experiments, as described in Section~\ref{subsubsec-online-success-rate}, using this trained model. Our findings, as illustrated in Fig. \ref{fig:ablation-study}, indicate that the model trained without our advanced augmentation method exhibits a notably high false positive rate in the subsequent online gripping task when compared to the results shown in Fig. \ref{fig:online-confusion} where the model was trained using our advanced augmentation method. In other words, the model trained without our advanced augmentation is unable to effectively distinguish patterns between slip and stop events. As a result, it incorrectly detects incipient slip in many stop events.


\subsection{Grasp stabilization after incipient slip detection}
This experiment aims to show the benefit of using our incipient slip detection method in practical gripping tasks. This involve lifting the robot arm while gripping the object with a pre-determined small force to ensure that slip occurs. We applied our incipient slip detection method and adjusted the grip when incipient slip was first detected to prevent the object from slipping further. In this experiment, we simulate two common scenarios that can trigger slips. The first involves gripping an object at its center of gravity with insufficient force and lifting it, causing a translational slip between the gripper and the object. The second involves gripping an object away from its center of gravity and lifting it, where rotational slip is likely to occur. We implemented a simple grip force adaptation that responds to incipient slip detection as follows: if incipient slip is detected, the robot immediately stops, and the gripper applies a pre-determined secure force to the object. The objects used in the experiment are the same as those shown in Fig.~\ref{fig:objects}. The experiment was conducted 36 times (6 objects $\times$ 2 scenarios (translation or rotation) $\times$ 3 repetitions). We fix ArUco markers on the objects and gripper and use Python OpenCV to track the positions and orientations of all.

We report the results in Table~\ref{table:online-results}, which demonstrate the quickly and effective detection of incipient slip using our algorithm. On average, our algorithm can timely detect incipient slip and prevent the object from slipping when the relative translation between the object and the gripper reaches 2.5~mm and the relative rotation reaches 1.9~$^{\circ}$. Our algorithm showcases its ability to facilitate timely corrective action, preventing object falls; a demonstration video can be seen at our project website given in the abstract.


\begin{table}[]
\centering
\caption{The translation and angular displacement between the object and the gripper after being lifted in our online gripping and lifting experiment, utilizing our incipient slip detection and grip correction methods.}
\begin{tabular}{l||cc}
Object         & Translation (mm)     & Rotation ($^{\circ}$)      \\ \hline\hline
Hard paper box & $4.5 \pm 0.6$        & $1.9 \pm 0.6    $   \\
Wooden box       & $2.4 \pm 0.8      $  & $1.5 \pm 0.9     $  \\
Shampoo bottle & $2.8  \pm 1.2      $ & $2.1 \pm 0.4    $   \\
Coffee jar     & $2.1 \pm 0.5      $  & $1.6 \pm 0.2    $   \\
Pringles can   & $1.4  \pm 0.4     $  & $1.1    \pm 0.3   $ \\
Thermos flask       & $2.1 \pm 0.6     $   & $1.2 \pm 0.4    $   \\ \hline\hline
Average        & $2.5 \pm 0.7 $                 & $1.9  \pm 0.5   $             
\end{tabular}
\label{table:online-results}
\vspace{-1.0em}
\end{table}
 
\section{Discussion} \label{Discussion}
Our developed algorithm enables the NN to effectively learn the incipient slip pattern from offline data and demonstrates high accuracy in both offline and online test sets. Furthermore, our algorithm enhances the security of robotic gripping.

Compared to previous related works \cite{papillarray-incipient-heba, papillarray-incipient-pablo}, our algorithm offers several advantages. Firstly, our incipient slip detection algorithm incorporates a data-driven learning-based approach, minimizing the need for extensive human involvement in investigating the complex patterns of incipient slip. Secondly, the improved robustness of our algorithm enables the NN to effectively adapt to diverse domains with various types of PapillArray sensors and robotic gripping systems, despite being trained solely on data lacking heterogeneity. Therefore, our algorithm is more practical and possesses greater potential for maximizing the utilization of valuable tactile data in real-world scenarios. Thirdly, our algorithm has the ability to distinguish between incipient slip and a closely related tactile pattern that we refer to as a stop event. Notably, previous related work \cite{papillarray-incipient-heba, papillarray-incipient-pablo, gross-slip-1, gelsight-detect-1} has not adequately considered or addressed the stop event; however, our investigation has revealed the importance of including stop events when developing incipient slip detection algorithms due to their similar patterns but entirely different consequences.




There are limitations to our work that need consideration. Firstly, the incipient slip detection could be improved by transitioning from a binary signal to a continuous warning signal. For instance, if incipient slip is detected in a small portion of the contact surface, the remaining area may still possess sufficient fraction to prevent significant slippage. In such cases, the warning level of incipient slip is low and corrective actions may not be necessary. Conversely, if a significant portion of the contact surface exhibits incipient slip, the warning level should escalate and it becomes important  to for appropriate corrective actions. Moreover, our current choice of force adaptation method for reacting to incipient slip falls short when compared to the state-of-the-art gripping control work \cite{papillarray-incipient-heba}. However, it is important to note that force adjustment is not the primary focus of our research in this paper, which is focused on improving the incipient slip detection. In future work, we will develop a more sophisticated force adaptation technique that incorporates our incipient slip detection method.

\section{Conclusion} \label{Conclusion}
In conclusion, this paper presents an incipient slip detection method that employs deep learning and several data augmentation techniques to improve the robustness of the trained NN. Our method is highly effective and reaches the state-of-art performance, it enable a single pre-trained NN model to be applied across various domains and tasks. In addition, our method has the potential to be extended to other approaches that use compliant tactile sensors.  

\appendix 
To train the NN parameters, we use stochastic gradient descent with a momentum of 0.95 and a learning rate of $10^{-3}$, with a batch size of 512. We also incorporate a weight decay of $10^{-3}$ using $L_{2}$ regularization during training. The encoder NN consists of one hidden layer with 1024 units, and the output dimension is 128. The GRU cell has a hidden layer dimension of 128. The predictor network comprises two hidden layers with 256 and 128 units, respectively. To all hidden layers, we apply rectified non-linearity \cite{relu} and batch normalization \cite{batch-norm}.

We implement our NN using PyTorch (Version 1.12.1, Meta, USA). All our experiments are conducted on a PC with an Intel 7-10875H CPU and an NVIDIA 2060 GPU. During the online evaluation stage, e utilise ROS \cite{ros} to facilitate communication between various components in our system.

\bibliographystyle{ieeetr}
\bibliography{main}

\begin{thebibliography}{10}

\bibitem{human-no-tactile}
R.~S. Johansson and G.~Westling, ``Roles of glabrous skin receptors and
  sensorimotor memory in automatic control of precision grip when lifting
  rougher or more slippery objects,'' {\em Experimental Brain Research},
  vol.~56, pp.~550--564, 1984.

\bibitem{human-no-tactile2}
A.-S. Augurelle, A.~M. Smith, T.~Lejeune, and J.-L. Thonnard, ``Importance of
  cutaneous feedback in maintaining a secure grip during manipulation of
  hand-held objects,'' {\em Journal of Neurophysiology}, vol.~89, no.~2,
  pp.~665--671, 2003.

\bibitem{human-hand-property}
A.~B. Vallbo, R.~S. Johansson, {\em et~al.}, ``Properties of cutaneous
  mechanoreceptors in the human hand related to touch sensation,'' {\em Hum
  Neurobiol}, vol.~3, no.~1, pp.~3--14, 1984.

\bibitem{incipient-slip-detection-survey}
W.~Chen, H.~Khamis, I.~Birznieks, N.~F. Lepora, and S.~J. Redmond, ``Tactile
  sensors for friction estimation and incipient slip detection—toward
  dexterous robotic manipulation: A review,'' {\em IEEE Sensors Journal},
  vol.~18, no.~22, pp.~9049--9064, 2018.

\bibitem{human-hand}
B.~P. Delhaye, E.~Jarocka, A.~Barrea, J.-L. Thonnard, B.~Edin, and P.~Lefevre,
  ``High-resolution imaging of skin deformation shows that afferents from human
  fingertips signal slip onset,'' {\em Elife}, vol.~10, p.~e64679, 2021.

\bibitem{papillarray-incipient-heba}
H.~Khamis, B.~Xia, and S.~J. Redmond, ``Real-time friction estimation for grip
  force control,'' in {\em 2021 IEEE International Conference on Robotics and
  Automation (ICRA)}, pp.~1608--1614, IEEE, 2021.

\bibitem{gross-slip-1}
J.~W. James, N.~Pestell, and N.~F. Lepora, ``Slip detection with a biomimetic
  tactile sensor,'' {\em IEEE Robotics and Automation Letters}, vol.~3, no.~4,
  pp.~3340--3346, 2018.

\bibitem{gross-slip-2}
M.~Sch{\"o}pfer, C.~Sch{\"u}rmann, M.~Pardowitz, and H.~Ritter, ``Using a
  piezo-resistive tactile sensor for detection of incipient slippage,'' in {\em
  ISR 2010 (41st International Symposium on Robotics) and ROBOTIK 2010 (6th
  German Conference on Robotics)}, pp.~1--7, VDE, 2010.

\bibitem{finger1}
S.~du~Bois~de Dunilac, D.~C{\'o}rdova~Bulens, P.~Lef{\`e}vre, S.~J. Redmond,
  and B.~P. Delhaye, ``Biomechanics of the finger pad in response to torsion,''
  {\em Journal of the Royal Society Interface}, vol.~20, no.~201, p.~20220809,
  2023.

\bibitem{finger2}
B.~Delhaye, P.~Lefevre, and J.-L. Thonnard, ``Dynamics of fingertip contact
  during the onset of tangential slip,'' {\em Journal of The Royal Society
  Interface}, vol.~11, no.~100, p.~20140698, 2014.

\bibitem{gross-slip-3}
Z.~Su, K.~Hausman, Y.~Chebotar, A.~Molchanov, G.~E. Loeb, G.~S. Sukhatme, and
  S.~Schaal, ``Force estimation and slip detection/classification for grip
  control using a biomimetic tactile sensor,'' in {\em 2015 IEEE-RAS 15th
  International Conference on Humanoid Robots (Humanoids)}, pp.~297--303, IEEE,
  2015.

\bibitem{modify-tactip-incipient}
J.~W. James, S.~J. Redmond, and N.~F. Lepora, ``A biomimetic tactile
  fingerprint induces incipient slip,'' in {\em 2020 IEEE/RSJ International
  Conference on Intelligent Robots and Systems (IROS)}, pp.~9833--9839, IEEE,
  2020.

\bibitem{papillarray-incipient-pablo}
P.~M. Ulloa, D.~C. Bulens, and S.~J. Redmond, ``Incipient slip detection for
  rectilinear movements using the papillarray tactile sensor,'' in {\em 2022
  IEEE Sensors}, pp.~1--4, 2022.

\bibitem{rnn}
D.~E. Rumelhart, G.~E. Hinton, and R.~J. Williams, ``Learning internal
  representations by error propagation,'' tech. rep., California Univ San Diego
  La Jolla Inst for Cognitive Science, 1985.

\bibitem{svm}
B.~E. Boser, I.~M. Guyon, and V.~N. Vapnik, ``A training algorithm for optimal
  margin classifiers,'' in {\em Proceedings of the Fifth Annual Workshop on
  Computational Learning Theory}, pp.~144--152, 1992.

\bibitem{tactip}
C.~Chorley, C.~Melhuish, T.~Pipe, and J.~Rossiter, ``Development of a tactile
  sensor based on biologically inspired edge encoding,'' in {\em 2009
  International Conference on Advanced Robotics}, pp.~1--6, IEEE, 2009.

\bibitem{cnn}
Y.~LeCun, L.~Bottou, Y.~Bengio, and P.~Haffner, ``Gradient-based learning
  applied to document recognition,'' {\em Proceedings of the IEEE}, vol.~86,
  no.~11, pp.~2278--2324, 1998.

\bibitem{gelsight}
W.~Yuan, S.~Dong, and E.~H. Adelson, ``Gelsight: High-resolution robot tactile
  sensors for estimating geometry and force,'' {\em Sensors}, vol.~17, no.~12,
  p.~2762, 2017.

\bibitem{gelsight-detect-1}
W.~Yuan, R.~Li, M.~A. Srinivasan, and E.~H. Adelson, ``Measurement of shear and
  slip with a gelsight tactile sensor,'' in {\em 2015 IEEE International
  Conference on Robotics and Automation (ICRA)}, pp.~304--311, IEEE, 2015.

\bibitem{gelsight-detect-3}
S.~Dong, D.~Ma, E.~Donlon, and A.~Rodriguez, ``Maintaining grasps within
  slipping bounds by monitoring incipient slip,'' in {\em 2019 International
  Conference on Robotics and Automation (ICRA)}, pp.~3818--3824, IEEE, 2019.

\bibitem{papillarray-instrument}
H.~Khamis, B.~Xia, and S.~J. Redmond, ``A novel optical 3d force and
  displacement sensor--towards instrumenting the papillarray tactile sensor,''
  {\em Sensors and Actuators A: Physical}, vol.~291, pp.~174--187, 2019.

\bibitem{papillarray-concept}
H.~Khamis, R.~I. Albero, M.~Salerno, A.~S. Idil, A.~Loizou, and S.~J. Redmond,
  ``Papillarray: An incipient slip sensor for dexterous robotic or prosthetic
  manipulation--design and prototype validation,'' {\em Sensors and Actuators
  A: Physical}, vol.~270, pp.~195--204, 2018.

\bibitem{human-tactile-speed}
R.~S. Johansson and A.~B. Vallbo, ``Tactile sensibility in the human hand:
  relative and absolute densities of four types of mechanoreceptive units in
  glabrous skin.,'' {\em The Journal of Physiology}, vol.~286, no.~1,
  pp.~283--300, 1979.

\bibitem{gru}
J.~Chung, C.~Gulcehre, K.~Cho, and Y.~Bengio, ``Empirical evaluation of gated
  recurrent neural networks on sequence modeling,'' {\em arXiv preprint
  arXiv:1412.3555}, 2014.

\bibitem{relu}
X.~Glorot, A.~Bordes, and Y.~Bengio, ``Deep sparse rectifier neural networks,''
  in {\em Proceedings of the Fourteenth International Conference on Artificial
  Intelligence and Statistics}, pp.~315--323, JMLR Workshop and Conference
  Proceedings, 2011.

\bibitem{batch-norm}
S.~Ioffe and C.~Szegedy, ``Batch normalization: Accelerating deep network
  training by reducing internal covariate shift,'' in {\em International
  Conference on Machine Learning}, pp.~448--456, PMLR, 2015.

\bibitem{ros}
M.~Quigley, K.~Conley, B.~Gerkey, J.~Faust, T.~Foote, J.~Leibs, R.~Wheeler,
  A.~Y. Ng, {\em et~al.}, ``Ros: an open-source robot operating system,'' in
  {\em ICRA workshop on open source software}, vol.~3, p.~5, Kobe, Japan, 2009.

\end{thebibliography}

\end{document}